\documentclass[conference]{IEEEtran}
\IEEEoverridecommandlockouts
\usepackage{cite}
\usepackage{amsmath,amssymb,amsfonts}
\usepackage{algorithmic}
\usepackage{graphicx}
\usepackage{textcomp}
\usepackage{xcolor}

\usepackage{hyperref}
\usepackage{booktabs}

\def\BibTeX{{\rm B\kern-.05em{\sc i\kern-.025em b}\kern-.08em
    T\kern-.1667em\lower.7ex\hbox{E}\kern-.125emX}}
\begin{document}

\title{SWAT-NN: Simultaneous Weights and Architecture Training for Neural Networks in a Latent Space\\
}

\author{
\IEEEauthorblockN{Zitong Huang}
\IEEEauthorblockA{\textit{Electrical and Computer Engineering} \\
\textit{University of Southern California}\\
Los Angeles, CA, USA \\
chuang95@usc.edu}
\and
\IEEEauthorblockN{Mansooreh Montazerin}
\IEEEauthorblockA{\textit{Electrical and Computer Engineering} \\
\textit{University of Southern California}\\
Los Angeles, CA, USA \\
mmontaze@usc.edu}
\and
\IEEEauthorblockN{Ajitesh Srivastava}
\IEEEauthorblockA{\textit{Electrical and Computer Engineering} \\
\textit{University of Southern California}\\
Los Angeles, CA, USA \\
ajiteshs@usc.edu}
}

\maketitle

\begin{abstract}
Designing neural networks typically relies on manual trial and error or a neural architecture search (NAS) followed by weight training. The former is time-consuming and labor-intensive, while the latter often discretizes architecture search and weight optimization. In this paper, we propose a fundamentally different approach that simultaneously optimizes both the architecture and the weights of a neural network. Our framework first trains a universal multi-scale autoencoder that embeds both architectural and parametric information into a continuous latent space, where functionally similar neural networks are mapped closer together. Given a dataset, we then randomly initialize a point in the embedding space and update it via gradient descent to obtain the optimal neural network, jointly optimizing its structure and weights. The optimization process incorporates sparsity and compactness penalties to promote efficient models. Experiments on regression datasets demonstrate that our method effectively discovers sparse and compact neural networks with strong performance.
\end{abstract}


\section{Introduction}
Traditionally, designing high-performing neural networks relies on a labor intensive and trial-and-error approach, which generally consists of defining a fixed architecture, training the weights of the network, evaluating performance, and iteratively refining the architecture to achieve better results. This iterative cycle is time-consuming, computationally expensive, and highly dependent on expert intuition. 

To address this challenge, Neural Architecture Search (NAS) has emerged as a promising framework for automatically discovering effective neural network architectures \cite{ren2021comprehensive, elsken2019neural}. Existing NAS methods can be broadly classified into discrete and continuous search approaches. 
Early NAS methods predominantly employed Reinforcement Learning(RL)~\cite{zoph2016neural} and evolutionary algorithms~\cite{liu2021survey,stanley2002evolving}, which fall under the discrete search category. RL-based methods \cite{baker2016designing, zhong2018practical} formulate architecture design as a sequential decision-making process, where each component of the architecture is treated as an action, and the objective is to find a sequence of actions that maximizes performance. Alternatively, evolutionary algorithms \cite{stanley2009hypercube, yao1999evolving, chen2019renas} search by iteratively mutating and recombining candidate operations to evolve better-performing architectures. Despite their differences, both approaches operate over high-dimensional discrete representations and navigate a vast combinatorial search space defined by a predefined set of operations. In contrast to discrete strategies, continuous NAS methods~\cite{elsken2019neural, liu2018darts} have gained attention for their efficiency and ability to operate in a differentiable search space\cite{santra2021gradient, xie2018snas, cai2018proxylessnas}.
These methods create a continuous (e.g., softmax) version of the decision objective, enabling search via gradient descent. The architecture is typically modeled as a computational graph, where each edge carries a soft weight over a predefined set of candidate operations. Nevertheless, while continuous NAS methods significantly reduce search cost and enable differentiable optimization, they are often constrained by task-specific predictors or datasets, and typically search for architectures and weights in a decoupled or alternating manner.

In this work, we propose SWAT-NN, which jointly optimizes both the architecture and its corresponding weights to achieve high performance on given datasets. It is based on a multi-scale autoencoder framework, trained to embed functionally similar neural networks close to one another within a continuous latent space. The encoder learns to represent the full neural network -- including both its structure and parameters -- in an embedding space, from which multiple decoders generate functionally similar neural networks of varying depths. This embedding, once trained, enables gradient-based optimization (with respect to the embedding, instead of weights) by directly optimizing task-specific performance. Because the latent representation jointly encodes architecture and weights, our method performs unified optimization over both, without requiring separate architectural parameters or surrogate models. We evaluate our proposed approach in the context of multi-layer perceptrons (MLPs) with three different activation functions: sigmoid, tanh, and leaky ReLU, and apply it to the Continuous Optimization Benchmark Suite
from Neural Network Regression (CORNN) benchmark, which contains 54 regression datasets derived from diverse benchmark function datasets \cite{malan2022continuous}. Notably, the proposed approach diverges from conventional NAS paradigms by treating networks holistically as function approximators, rather than representing them as graphs composed of discrete operations. This allows the embedding to be over their actual functional behavior. Our main contributions can be summarized as follows:
\begin{enumerate}
    \item We propose a new framework, SWAT-NN, that performs simultaneous optimization over neural network architectures and weights within a universal embedding space, departing fundamentally from traditional NAS methods that decouple architecture search and weight tuning.
    \item We show that SWAT-NN supports fine-grained architecture discovery, including neuron-level activation function selection and layer-wise width adaptation.
    \item Through experiments on 54 regression datasets, we show that SWAT-NN discovers significantly sparser and more compact models compared to existing methods, while maintaining comparable or better accuracy.
\end{enumerate}

\section{Related Work}
Neural networks can be designed manually or through automated search techniques known as Neural Architecture Search (NAS). Traditional NAS approaches include reinforcement learning (RL)~\cite{zoph2016neural} or evolutionary algorithms~\cite{liu2021survey,stanley2002evolving}: RL formulates architecture design as a sequential decision-making process, whereas evolutionary methods construct architectures by iteratively mutating or recombining predefined operations. However, these methods typically operate in a discrete search space, making them computationally expensive and difficult to scale. More recent efforts explore continuous NAS, which relaxes the discrete architecture space into a continuous domain.

Differentiable Architecture Search (DARTS)~\cite{liu2018darts} is one of the most influential gradient-based NAS methods. It relaxes discrete architectural choices by introducing a softmax over all candidate operations on each edge of a directed acyclic graph (DAG). The search process is framed as a bi-level optimization problem, in which the network weights are optimized on the training set, while the architecture parameters are updated on the validation set. Such decoupling breaks joint optimization and can result in suboptimal architectures due to misaligned objectives between weight and architecture updates.

Neural Architecture Optimization (NAO) based on a graph variational autoencoder (VAE)~\cite{li2020ngae} encodes network architectures into a continuous latent representation specific to the given set of datasets. It uses a surrogate performance predictor, a regression model estimating task-specific accuracy of an architecture based on its latent representation, to guide the search. Specifically, it performs gradient ascent in the latent space to find architectures that are predicted to perform well for the given dataset. While NAO improves the smoothness of the optimization landscape and captures mathematical properties of architectures through its encoder, it fully decouples architecture discovery from weight optimization and relies on dataset-specific predictors. Further, it requires retraining of the autoencoder from scratch for each new set of tasks.

In contrast, we propose a novel approach that departs fundamentally from conventional NAS formulations. Rather than searching over discrete or graph-based compositions of operations, we treat entire neural networks as function approximators and embed their complete mathematical behavior -- including both architecture and weights -- into a universal continuous latent space. Optimization is performed directly on this functional representation, without relying on manually defined search spaces or architectural priors. Moreover, our method is agnostic to specific datasets, as it does not depend on task-specific performance predictors or surrogate models, enabling more general and flexible network discovery. Unlike many the existing methods, which generate the networks through a sequence of decisions~\cite{baker2016designing, zhong2018practical,li2020ngae,liu2021survey,stanley2002evolving}, our approach decodes a complete neural network from a point in the embedding space.

\section{Methodology}
\begin{figure}[h]
  \centering
  \includegraphics[width=\linewidth]{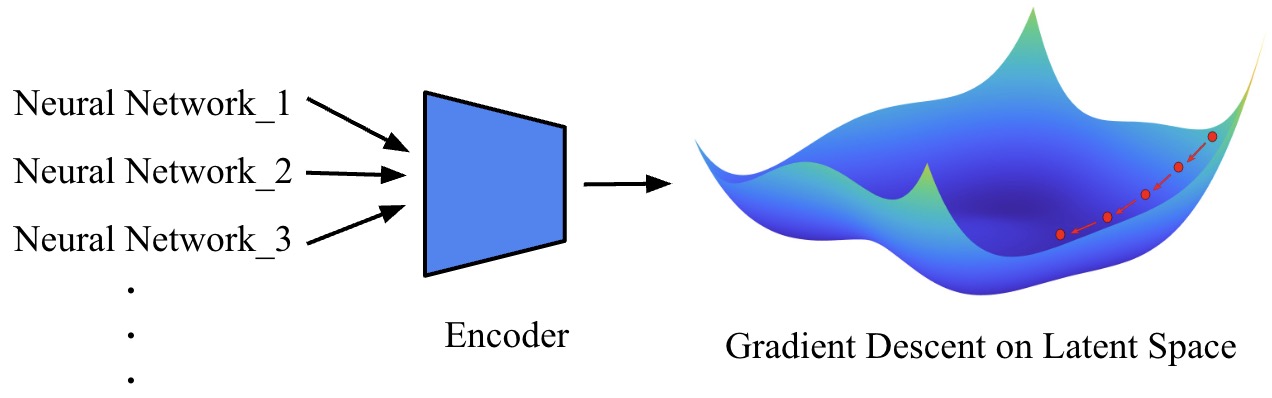}
  \caption{An illustration of the overall idea for SWAT-NN}
  \label{pipeline}
\end{figure}

 The key idea of SWAT-NN is based on achieving the following two goals (see \autoref{pipeline}): (1) training a multi-scale autoencoder to construct a latent embedding space for multi-layer perceptrons (MLPs) (\autoref{Embedding MLPs using Autoencoder}); and (2) searching for sparse and compact networks via gradient descent in the learned embedding space (\autoref{Training MLP in the Embedding Space}). 
This paper focuses on training MLPs with a varying number of neurons per layer and different activation functions per neuron.

\subsection{Embedding MLPs with Autoencoder}
\label{Embedding MLPs using Autoencoder}
We aim to encode both the architecture and corresponding weights of a neural network into a shared embedding space. This is achieved through a multi-scale autoencoder framework. The encoder, denoted as $E$, maps a given MLP into a latent representation. A set of $L$ decoders, denoted as ${D_1, D_2, \dots, D_L}$, then, generate MLPs with 1 to $L$ hidden layers, respectively. In this work, we set $L = 4$. The autoencoder is trained to generate MLPs that may not be reconstructions of the input networks, but are functionally similar -- that is, they produce similar outputs when given the same inputs. In the following, we provide further details on the MLP representation and autoencoder design:

\subsubsection{Matrix Representation of MLPs}
\label{Matrix Representation of MLPs}
To train an autoencoder for encoding neural networks, we require a structured and fixed-size representation of MLPs that can act as input. Both the architecture and weights of an MLP are encoded in a concise matrix-based representation. For an MLP with $L$ hidden layers, each containing $N$ neurons, input dimension $i$, and output dimension $o$, the weights between consecutive layers can be expressed as a sequence of matrices:
\begin{equation}
\label{matrix_repr}
    [(W_{i,N}, b_{N}), (W_{N,N}, b_{N}), \dots, (W_{N,N}, b_{N}), (W_{N,o}, b_{o})]
\end{equation}
where $W_{i,N} \in \mathbb{R}^{i \times N}, W_{N,N} \in \mathbb{R}^{N \times N}$, and $W_{N,o} \in \mathbb{R}^{N \times o}$ are weight matrices between layers, and the corresponding bias vectors are $b_N \in \mathbb{R}^{N \times 1}$, $b_o \in \mathbb{R}^{o \times 1}$. To ensure uniform dimensionality for concatenation, we apply zero-padding to the first and last matrices so that all matrices have dimensions of $N \times N$. In addition, each bias vector is padded to $N \times 1$ and horizontally concatenated into the matrix of weights. To differentiate between the actual weights and padded entries, a secondary mask matrix is appended in parallel to the matrix described above. More specifically, each position corresponding to an original MLP weight is assigned a value of 1 in the mask, while positions introduced through zero-padding are assigned a value of 0. The sizes of the input and output layers can also be flexibly specified using boundary-layer masks (see \autoref{disc:compression}). This basic representation is later extended to incorporate varying numbers of neurons and activation functions in \autoref{Varying Number of Neurons} and \autoref{Changing Activation Functions}; an overview of the complete representation scheme is illustrated in \autoref{fig:matrix_repr}.

\subsubsection{Multi-scale Autoencoder}
\label{Multi-scale Autoencoder}
\begin{figure*}
  \centering
  \includegraphics[width=\linewidth]{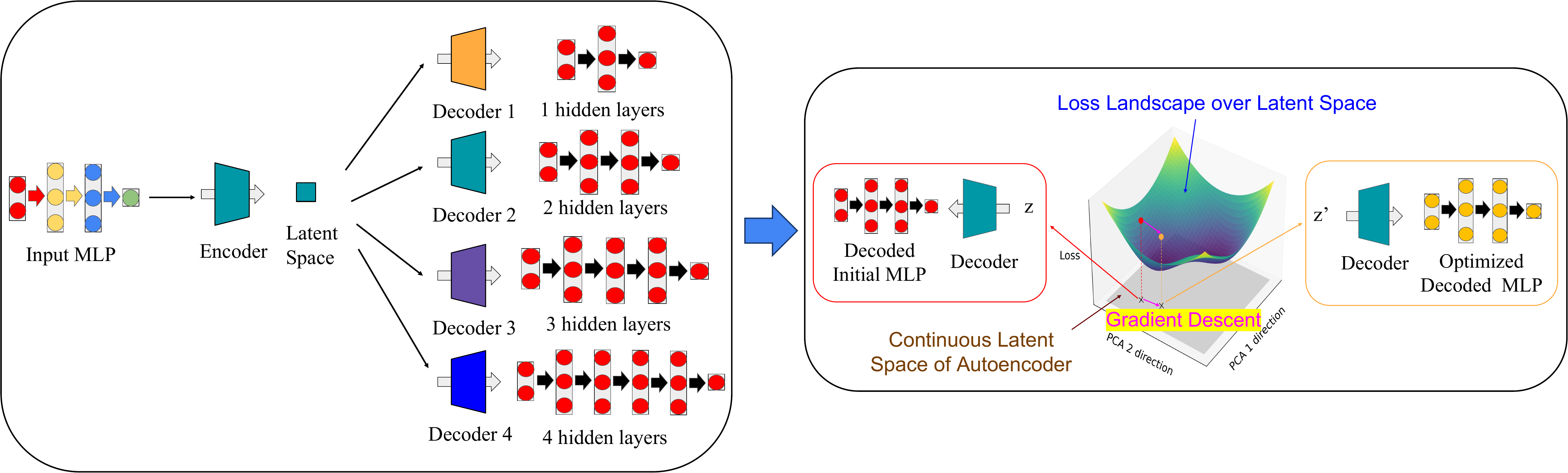}
  \caption{Left: Multi-scale autoencoder architecture with four decoders, each corresponding to MLPs with 1–4 hidden layers. Right: Pipeline for training optimal MLPs through gradient-based search in the continuous latent space learned by the autoencoder.}
  \label{autoencoder}
\end{figure*}

The multi-scale autoencoder consists of a single encoder $E$ and decoders $D_1$, $D_2$, $\dots$, $D_L$, where each decoder $D_k$ generates an MLP with $k$ hidden layers. The overall structure is illustrated in \autoref{autoencoder}.

The encoder and decoders are implemented using a GPT-2 architecture~\cite{radford2019language}. Following the matrix representation of MLPs described in \autoref{Matrix Representation of MLPs}, each row $h$ corresponds to the concatenation of all outgoing weights from the 
$h$\textsuperscript{th} neuron across every layer of the MLP. In other words, the $h$\textsuperscript{th} row aggregates the weights from the $h$\textsuperscript{th} neuron in each layer to all neurons in the subsequent layers, providing a unified view of its outgoing connections throughout the network. Accordingly, we treat the set of outgoing weights from the $h$\textsuperscript{th} neuron across all layers as a single unified token. Each row $h$ is, then, linearly projected to be compatible with the token embedding size of the GPT-2 encoder and decoders. This representation bounds the number of tokens by the maximum number of neurons per layer, thereby minimizing computational cost. By concatenating outgoing weights across layers, the model is further enabled to capture long-range dependencies. 

During the encoding stage, the MLP matrix representation is provided as input to the encoder, which maps it to a latent embedding. This embedding is then passed to $L$ decoders, each corresponding to a different target MLP depth. In the decoding stage, each decoder uses a GPT-2-style autoregressive generation process to produce a new matrix representation of an MLP. For decoder $D_k$, we retain only the first $k$ hidden-layer blocks from the generated output matrix, yielding the final representation of an MLP with $k$ hidden layers.

\subsubsection{Varying Number of Neurons}
\label{Varying Number of Neurons}
To enable the autoencoder to account for different numbers of neurons per layer as part of the network architecture, we introduce a neuron indicator matrix to explicitly encode this structural variation. For an MLP with $L$ hidden layers, each with at most $N$ neurons, we augment the matrix representation introduced in \autoref{Multi-scale Autoencoder} by appending an additional mask matrix $M \in \{0,1\}^{N \times L}$. Each column of $M$ corresponds to one hidden layer, which indicates the activation status of all the neurons in that layer. Specifically, $M_{h,j} = 1$ if the $h$th neuron in the $j$th hidden layer is active in the input MLP, and $M_{h,j} = 0$ otherwise. 

We apply the autoencoder introduced in \autoref{Multi-scale Autoencoder} to this extended representation. The decoded neuron indicator matrix $\hat{M} \in [0,1]^{N \times L}$ is obtained via a sigmoid activation, where each element $\hat{M}_{h,j}$ represents the probability of the $h$th neuron in the $j$th hidden layer being active. To preserve the differentiability of the loss function during training, we avoid hard thresholding for enforcing binary activations. Instead, we use a soft thresholding method to decide entirely active vs pruned neurons described in \autoref{Sparsity and Compactness}.

\subsubsection{Changing Activation Functions}
\label{Changing Activation Functions}
\begin{figure*}
    \centering
    \includegraphics[width=0.8\linewidth]{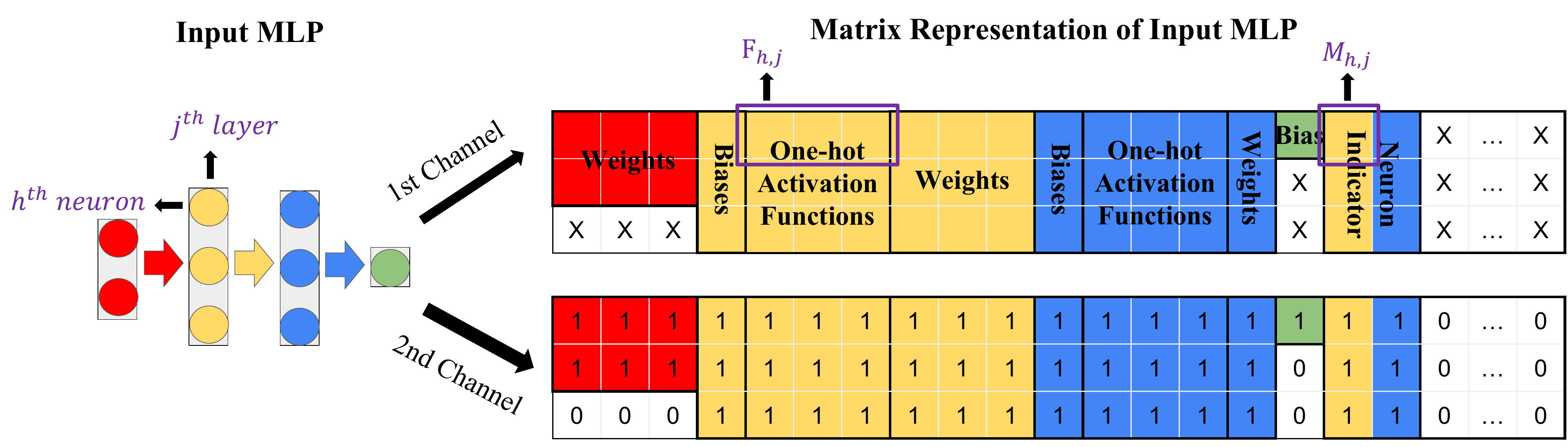}
    \caption{Matrix representation for a 2 hidden-layer MLP with 3 neurons per layer. Different colors indicate elements associated with neurons from different layers, encoding the outgoing weights, activation functions, and bias terms of each neuron. Entries marked with 'X' represent zero-padding.}
    \label{fig:matrix_repr}
\end{figure*}

To capture activation function choices as part of the network architecture, we extend the matrix representation to explicitly encode which activation function is applied to each neuron. We incorporate activation function choices into the matrix representation introduced in \autoref{Matrix Representation of MLPs} and \autoref{Varying Number of Neurons} through appending additional columns that specify the activation function used by each neuron. Assuming a total of $A$ possible activation functions, the activation type of each neuron is encoded as a one-hot vector of length $A$. The final representation is:
\begin{align}
    [ & (W_{i,N}, b_{N}),\ F_{N,A},\ (W_{N,N}, b_{N}),\ F_{N,A}, \nonumber \\
      & \dots,\ (W_{N,N}, b_{N}),\ F_{N,A},\ (W_{N,o}, b_{o}),\ M_{N,L} ]
\end{align}
The final matrix $M_{N,L} \in \{0,1\}^{N \times L}$ indicates neuron activity as described in \autoref{Varying Number of Neurons}. Each $F_{N,A} \in \{0,1\}^{N \times A}$ follows a hidden layer’s weight and bias matrix and encodes the activation function for each neuron in that layer. An illustration of the whole matrix representation is shown in \autoref{fig:matrix_repr}.

After feeding the extended representation into the encoder-decoder framework, we apply continuous relaxation to decoded $\hat{F}_{N,A}$ matrices to enable differentiable selection of activation functions. Each row of $\hat{F}_{N,A}$ is interpreted as a softmax distribution over the $A$ candidate activation functions. Formally, for a neuron $h$ in layer $j$, the output is computed as a soft combination of candidate activation functions:
\begin{equation}
    y_{(h,j)}(x) = \sum_{k=1}^{A} \alpha_{(h, j),k} \cdot o_k(x)
\end{equation}
where $x$ is the input flows through current neuron $(h,j)$, and $o_k(\cdot) \in \mathcal{O}$ denotes the $k$th candidate activation function from a predefined activation function set $\mathcal{O}$. The activations' selection probabilities are given by:
\begin{equation}
    \alpha_{(h,j),k} = \text{softmax}(\hat{F}^{(j)}_{h,:})_k
\end{equation}
Here, $\hat{F}^{(j)}_{h,:} \in \mathbb{R}^A$ denotes the decoded activation logits for the $h$\textsuperscript{th} neuron in the $j$\textsuperscript{th} hidden layer. This relaxation preserves differentiability during training, enabling smooth optimization over activation selections. A near-discrete activation assignment can be enforced at inference time in \autoref{Activation Selection}.

\subsubsection{Training Autoencoder}
\label{Training Autoencoder}
We generate input vectors $x$ , and feed them into the input MLP $\mathcal{N}_s$ to obtain the corresponding outputs. Each input MLP is then passed through the autoencoder to produce decoded MLPs with varying depths $\{1, 2, \dots, L\}$. To capture the diverse functional properties of neural networks, we train the autoencoder on a large-scale collection of synthetically generated MLPs spanning a wide range of architectures and behaviors.
We adopt a minimum-loss objective to encourage functional similarity between the input MLP and its decoded variants. The loss function is defined as:
\begin{equation}
L_{ED} = \sum_{\mathcal{N}_s} \min_{i \in {1, 2, \dots, L}} \left( \sum_{x} \left( \mathcal{N}_s(x) - [D_i(E(\mathcal{N}_s))](x) \right)^2 \right)
\label{min loss}
\end{equation}
where $\mathcal{N}_s(x)$ denotes the output of the original input MLP evaluated at input $x$, and $[D_i(E(\mathcal{N}_s))](x)$ denotes the output of the decoded MLP with $i$ hidden layers, reconstructed from the bottleneck embedding and evaluated at the same input $x$. This objective encourages at least one decoded MLP to closely approximate the functional behavior of the original MLP across all sampled inputs. It effectively pulls functionally similar MLPs closer together in the embedding space.

The runtime complexity for training a single encoder and $L$ decoders is $\mathcal{O}\left( E \cdot \frac{N}{B} \cdot L \cdot f \right)$, where $f$ is the cost of a forward or backward pass through GPT-2, $N$ is the dataset size, $B$ is the batch size, and $E$ is the number of training epochs.

\subsection{Training MLP in the Embedding Space}
\label{Training MLP in the Embedding Space}
Given a dataset $S$, we leverage the multi-scale autoencoder, where functionally similar MLPs are smoothly embedded close to each other, to search for a sparse and compact MLP tailored to $S$. The decoded MLP can vary in the number of hidden layers (up to $L$), the number of neurons per layer (up to $N$), sparsity patterns, and activation functions across neurons (selected from $\mathcal{O}$). All architectures lie within the structural constraints defined during autoencoder training.

\subsubsection{Training Optimal MLPs for a Dataset}
\label{Training Optimal MLPs for a Dataset}
The learnable parameters of the multi-scale autoencoder described above are fixed during this stage. Since all MLPs are embedded into a continuous latent space, the problem of finding an optimal MLP reduces to optimizing an embedding vector $z \in \mathbb{R}^d$ within this space. Given a dataset $S = \{(x, y)\}$, we define the objective function for decoder $D_i$ as:
\begin{equation}
\label{original loss}
L_i(z) = \sum_{(x, y) \in S} \left( [D_i(z)](x) - y \right)^2
\end{equation}
where $[D_i(z)](x)$ denotes the output of the decoded MLP with $i$ hidden layers evaluated at input data $x$. The gradient of $L_i$ with respect to $z$ can be written as:
\begin{equation}
\nabla_z L_i = 2 \sum_{(x, y) \in S} \left( [D_i(z)](x) - y \right) \nabla_z [D_i(z)](x)
\end{equation}

To solve this optimization problem, we initialize $z$ by randomly sampling an embedding vector, and performing gradient descent to minimize $L_i(z)$ with respect to $z$. Specifically, the update rule can be defined as:
\begin{equation}
z^{(t+1)} = z^{(t)} - \eta \frac{\partial L_i}{\partial z^{(t)}}
\end{equation}
where $\eta$ is the learning rate and $t \in \{0, 1, \dots, T\}$ denotes the optimization step. The runtime complexity for this optimizing is
$\mathcal{O}\left( L \cdot T \cdot f \right)$,
where $L$ is the number of decoders, $T$ is the number of gradient steps per decoder, and $f$ is the cost of a forward or backward pass through a GPT-2 decoder.

\subsubsection{Sparsity and Compactness}
\label{Sparsity and Compactness}
To promote neural networks that are both sparse (fewer connections) and compact (fewer active neurons), we incorporate two regularization terms into the loss: a weight sparsity penalty $\mathcal{P}_{\text{s}}$ to drive individual weights toward zero, and a neuron compactness penalty $\mathcal{P}_{\text{n}}$ to suppress redundant neurons and promote compact architectures. The overall loss function is defined as:
\begin{align}
\label{eq:overall_sp_loss}
L_i (z) &= \sum_{(x, y) \in S} \left( [D_i(z)](x) - y \right)^2  + \lambda_s \cdot \mathcal{P}_{\text{s} } + \mathcal{P}_{\text{n}}
\end{align}
where $\mathcal{P}_{\text{s}}$ denotes the sparsity penalty weighted by coefficients $\lambda_s$, and $\mathcal{P}_{\text{n}}$ denotes the neuron compactness penalty.

\textbf{Sparsity Penalty.} $\mathcal{P}_{\text{s}}$ consists of two components: an $\ell_1$-regularization term and a soft counting switch. Formally, it is defined as:
\begin{align}
\label{eq:sp_loss}
    \mathcal{P}_{\text{s}} &= \mathcal{L}_1 + \text{SoftCount}\,, \\
    \mbox{Where, }
    \mathcal{L}_1 &= \mu_1 ||W[D_i(z)]||_1\,, \\ 
    \mbox{And, }
    \text{SoftCount} &= \mu_c \cdot \sigma \left( 20 \cdot \left\||W[D_i(z)]| - t_s \right\|_1 \right)
\end{align}
Here, $W[D_i(z)]$ denotes the weight matrix of the decoded MLP $D_i(z)$, and $t_s$ is a learnable threshold parameter. The $\mathcal{L}_1$ term encourages weights to shrink toward zero. The soft counting switch approximates the number of weights below the threshold $t_s$, using a scaled sigmoid function to softly suppress small-magnitude weights. During training, weights smaller than $t_s$ are softly masked using a sigmoid-based function that smoothly pushes them toward zero; at test time, they are hard-masked. Both $\mu_1$ and $\mu_c$ are tunable hyperparameters that control the relative strength of $\ell_1$ regularization and the soft counting penalty, respectively.

\textbf{Neuron Compactness Penalty.} To reduce the number of active neurons per layer by pruning unnecessary ones, we define the neuron compactness penalty $\mathcal{P}_{\text{n}}$ as a combination of two terms: the negative variance and the average magnitude of soft activation masks across neurons within each layer:
\begin{equation}
\label{eq:compact_loss}
\mathcal{P}_{\text{n}} = - \alpha \cdot\frac{1}{L} \sum_{i=1}^{L} \text{std}(M_i) + \beta \cdot \frac{1}{L} \sum_{i=1}^{L} \text{mean}(M_i)
\end{equation}
Here, $M_i$ denotes the soft neuron activation indicators for the $i$\textsuperscript{th} hidden layer, and $L$ is the number of layers. The first term encourages the activation probabilities within each layer to become more polarized, 
while the second term penalizes the average activation level to further suppress unnecessary neurons. Coefficients $\alpha$ and $\beta$ are hyperparameters controlling the strength of each term. During training, neuron indicators smaller than a fixed threshold $t_n = 0.5$ are softly masked using a sharpened sigmoid function that closely approximates hard thresholding while maintaining differentiability; at test time, neurons this threshold are hard-masked as inactive. 


\subsubsection{Activation Function Selection}
\label{Activation Selection}
We adopt a temperature-controlled softmax mechanism \cite{liu2018darts} to gradually fix the activation function assigned to each neuron. We define a temperature schedule over training epochs as follows:
\begin{equation}
    T(e) = \max\left(T_{\text{final}}, T_{\text{init}} \cdot \left(1 - \frac{e}{E_{\text{anneal}}} \right)\right)
\end{equation}
Here, $T_{\text{init}}$ is the initial temperature at the beginning of training, $T_{\text{final}}$ is the minimum temperature reached. $E_{\text{anneal}}$ is the number of epochs over which the temperature linearly decays to $T_{\text{final}}$, and $e$ denotes the current training epoch.

During training, for each neuron, a softmax over activation function candidates (i.e., ReLU, Tanh, Sigmoid) is applied using the current temperature $T(e)$. Let $\mathbf{a}_{h,j} \in \mathbb{R}^A$ denote the unnormalized logits for $A$ activation candidates of the $h$\textsuperscript{th} neuron in layer $j$; the contribution weight of activation operation $a$ at neuron $(h,j)$ is computed as:
\begin{equation}
    \alpha_{\text{h,j,a}} = \frac{\exp\left(\mathbf{a}_{h,j,a} / T(e)\right)}{\sum\limits_{a=1}^{A} \exp\left(\mathbf{a}_{h,j,a} / T(e)\right)}
\end{equation}

As training progresses and $T(e)$ decreases, the softmax distribution sharpens, eventually approximating a one-hot selection. At test time, we select the activation function with the highest softmax score for each neuron as its final activation.

\section{Experiments and Results}
\label{all_exp}

Our experiments are structured into three components:
\begin{itemize}
    \item \textbf{Exp 1: Learning Sparse and Compact MLPs (Joint Optimization of Activation, Neuron Usage, and Sparsity).}  
    We evaluate SWAT-NN's ability to jointly optimizes activation functions, neuron usage, and weight sparsity. We compare against DARTS, a baseline that supports differentiable architecture search, followed by Alternating Direction Method of Multipliers (ADMM) pruning.
    \item \textbf{Exp 2: Search for Activation Functions.}  
    To isolate the effect of neuron-level activation function selection, we evaluate our framework described in \autoref{Activation Selection} and compare it against two baselines: (a) autoencoder-based search with fixed activation functions, and (b) traditional MLP training with fixed activation functions.
    \item \textbf{Exp 3: Search for Sparse MLPs.}  
    To further isolate the effect of weight sparsity, we fix neuron counts and optimize for sparse weights, as described in \autoref{Sparsity and Compactness}. We compare against a baseline where autoencoder-based search is followed by ADMM pruning.
\end{itemize}

All the code is publicly available~\footnote{\url{https://github.com/zitonghuangcynthia/SWAT-NN_code}}.
\subsubsection{Dataset}
All evaluations are conducted on the CORNN benchmark suite~\cite{malan2022continuous}, which consists of 54 regression datasets. Each dataset defines a distinct function $f: \mathbb{R}^2 \rightarrow \mathbb{R}$, designed to capture a wide range of functional behaviors, including periodicity, discontinuities, sharp curvature, and varying smoothness. For each dataset, the data is split into 3,750 training samples and 1,250 test samples. All inputs are normalized to lie within $[-1, 1]^2$, and outputs are scaled to the range $(-1, 1)$.

\subsubsection{Autoencoder Settings}
We trained a multi-scale GPT-2 autoencoder with one encoder and four decoders for MLPs with 1-4 hidden layers. Each input MLP has two inputs, one output, and up to 7 neurons per hidden layer. Weights and biases are sampled from $[-5, 5]$ and $[-1, 1]$, respectively. For each MLP, 1000 inputs are uniformly sampled from $[-1, 1]^2$, and the autoencoder is trained using the loss function in \autoref{Training Autoencoder}. Each training epoch includes 50,000 batches (batch size 64) and takes around 3 hours to complete on a single RTX A5000 GPU. In \autoref{disc:compression}, we also incorporate varying input and output dimensions by applying the same masking and padding strategy to the boundary layers.

\subsubsection{Baselines}
\label{Baselines}
Most NAS methods do not perform truly simultaneous optimization of both architecture and weights. Instead, they decouple the process and are dataset-specific. In contrast, our method jointly optimizes structure and parameters in a universal embedding space, enabling direct gradient-based optimization for sparse and performant neural networks.

\textbf{Exp 1: DARTS with ADMM} 
Although no prior work jointly optimizes both architecture and weights, we compare against a baseline, evaluated in \autoref{exp:all_components}, that adapts DARTS to MLPs, allowing differentiable search over number of neurons per layer and neuron-level activation functions. The resulting architecture is then pruned using ADMM, an optimization method that alternates between minimizing loss and enforcing sparsity via auxiliary variables and dual updates.

The DARTS search space is a directed acyclic graph (DAG), where each node represents a layer-wise feature, and each edge encodes a candidate layer configuration defined by neuron count and activation functions. Softmax relaxation enables gradient-based optimization of both. Unlike the original DARTS designed for CNNs or RNNs, we omit cell repetition and reduction layers, as these are not applicable to fully connected MLPs without spatial or sequential hierarchies.

We train the DARTS model for 50 epochs to search for the architecture using bilevel optimization\cite{liu2018darts} procedure. ADMM is then applied with $\rho = 2$ and a threshold of $10^{-1}$ to sparsify the weights. These hyperparameters are selected after increasing sparsity strength until performance on roughly one-third of the tasks dropped over 5\% -- our tolerable limit.

\textbf{Exp 2: Autoencoder-based Search with Fixed Activation Functions} 
Used as a baseline in \autoref{Search for Optimal MLPs with Activation Functions}, we fix the activation function for all neurons to one of Sigmoid, Tanh, or Leaky ReLU, and use the trained autoencoder to optimize over weights (8000 epochs, learning rate 0.1). 

\textbf{Exp 2: Traditional MLP Training} 
Also evaluated in \autoref{Search for Optimal MLPs with Activation Functions}, we train MLPs with fixed activation functions and randomly initialized weights using standard gradient descent for 6000 epochs at a learning rate of 0.01. This reflects a common trial-and-error approach.

\textbf{Exp 3: Autoencoder with ADMM} 
Used as a baseline in \autoref{Search for Sparse and Compact MLPs}, we combine autoencoder-based architecture and weight optimization without sparsity penalty with post-hoc ADMM pruning. Models are trained for 8000 epochs with a learning rate of 0.1, while the ADMM uses $\rho = 2$ and a pruning threshold of $10^{-1}$.

\subsection{Learning Sparse and Compact MLPs
(Joint Optimization of Activation, Neuron Usage, and Sparsity)}
\label{exp:all_components}

In SWAT-NN setting, for sparsity penalty, we set $\lambda_s$ to $1 \times 10^{-4}$, $\mu_1 = 0.1$ for the $\ell_1$ penalty and $\mu_c = 0.01$ for the soft count term. For neuron compactness penalty, we use $\alpha = 0.4$ and $\beta = 0.001$. For activation function selection, we set $T_{\text{init}}$ = 1.0, $T_{\text{final}}$ = 0.01, $E_{\text{anneal}}$ = 3000. These hyperparameters are tuned using the last two benchmark functions (F53 and F54) from the CORNN benchmark suite as a validation set. SWAT-NN training takes about 1 minute per task and per MLP configuration on a machine with a single RTX A5000 GPU, 128GB RAM, and a 56-core CPU.

\begin{figure}
    \centering
    \includegraphics[width=1\linewidth]{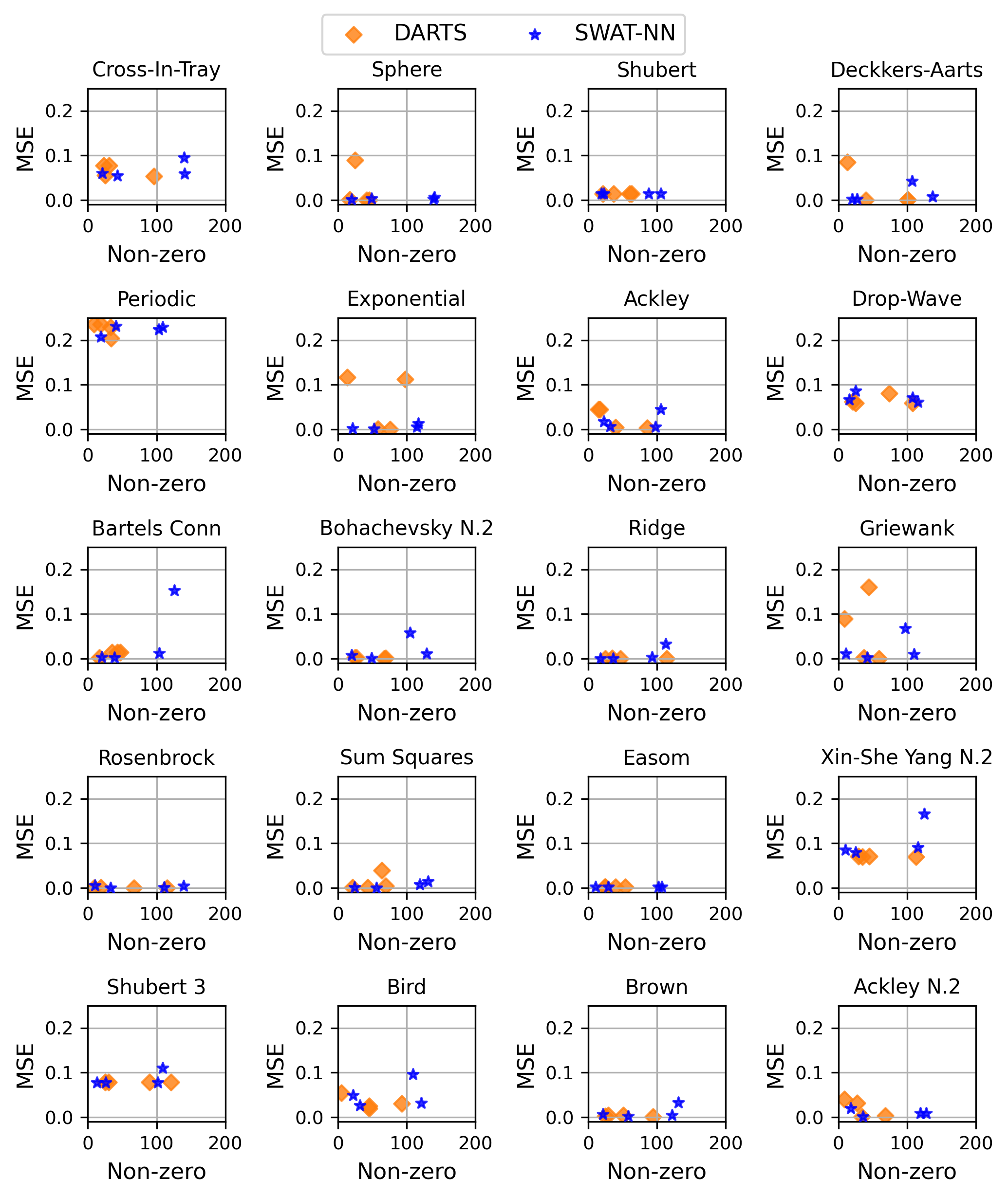}
    \caption{MSE vs non-zero weights between DARTS+ADMM and SWAT-NN across 20 randomly selected datasets. Each point represents a model configuration, plotted by its test MSE and number of non-zero weights.}
    \label{fig:darts_pareto}
\end{figure}

 \autoref{fig:darts_pareto} presents MSE vs non-zero weights for 20 datasets, where each point represents a model configuration, and positions closer to the bottom-left indicate lower MSE and higher sparsity. SWAT-NN always identifies solutions closer to the optimal accuracy-sparsity trade-off, demonstrating its advantage in producing more compact and efficient networks.

\begin{figure}
    \centering
    \includegraphics[width=1\linewidth]{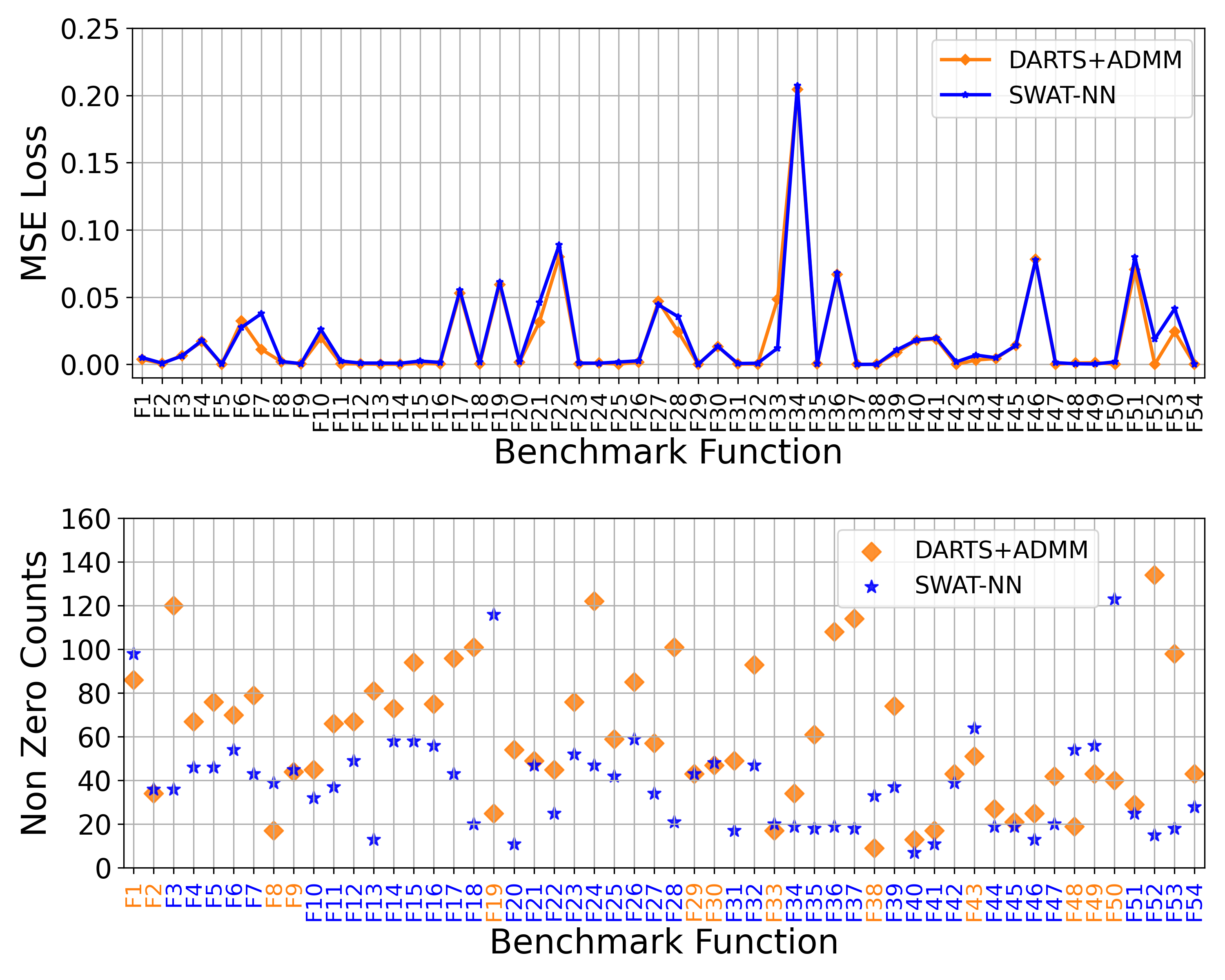}
    \caption{Top: Test MSE of the best-performing models identified by DARTS+ADMM and SWAT-NN across all 54 datasets. Bottom: Corresponding number of non-zero weights. Function labels are color-coded to indicate which method yields a more sparse and compact model.}
    \label{fig:darts_best_model}
\end{figure}

\autoref{fig:darts_best_model} show the results on all datasets in terms of MSE and number of non-zero weights. For each dataset, we apply a two-step selection criterion: we first identify architectures whose MSE is within 5\% of the minimum, and among them select the one with the fewest non-zero weights. While SWAT-NN and DARTS followed by ADMM achieve similar test accuracy, in most cases, SWAT-NN consistently discovers models with significantly fewer non-zero weights.

\subsection{Search for Optimal MLPs with Activation Functions}
\label{Search for Optimal MLPs with Activation Functions}
We focus exclusively on the search over activation functions. We use all 7 neurons in each hidden layer and do not apply the sparsity penalty discussed in \autoref{Sparsity and Compactness} during training. The same hyperparameters for activation function selection are used as described in the \autoref{exp:all_components} experiment. Each experiment is repeated three times, and we report the average MSE across runs.

\begin{figure}
    \centering
    \includegraphics[width=1\linewidth]{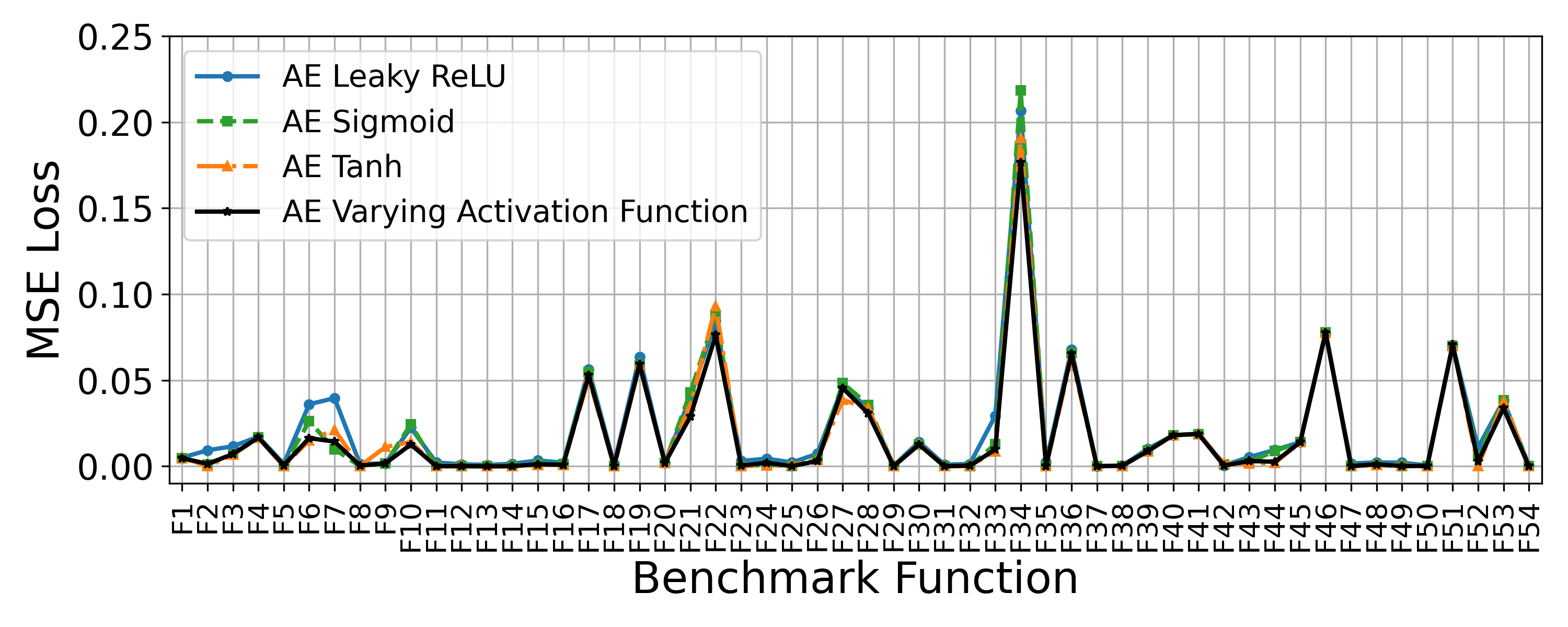}
    \caption{Comparison of MSE using autoencoder-based search with fixed activation functions versus neuron-level activation function search.}
    \label{fig:ae_act_noact}
\end{figure}

We compare our method, which searches neuron-level activation functions in the embedding space, with autoencoder-based search with fixed activation
functions described in \autoref{Baselines}. 
As shown in \autoref{fig:ae_act_noact}, our method consistently achieves performance comparable to or better than the best fixed-activation baselines, demonstrating that searching over activation types improves model performance.


\begin{figure}
    \centering
    \includegraphics[width=1\linewidth]{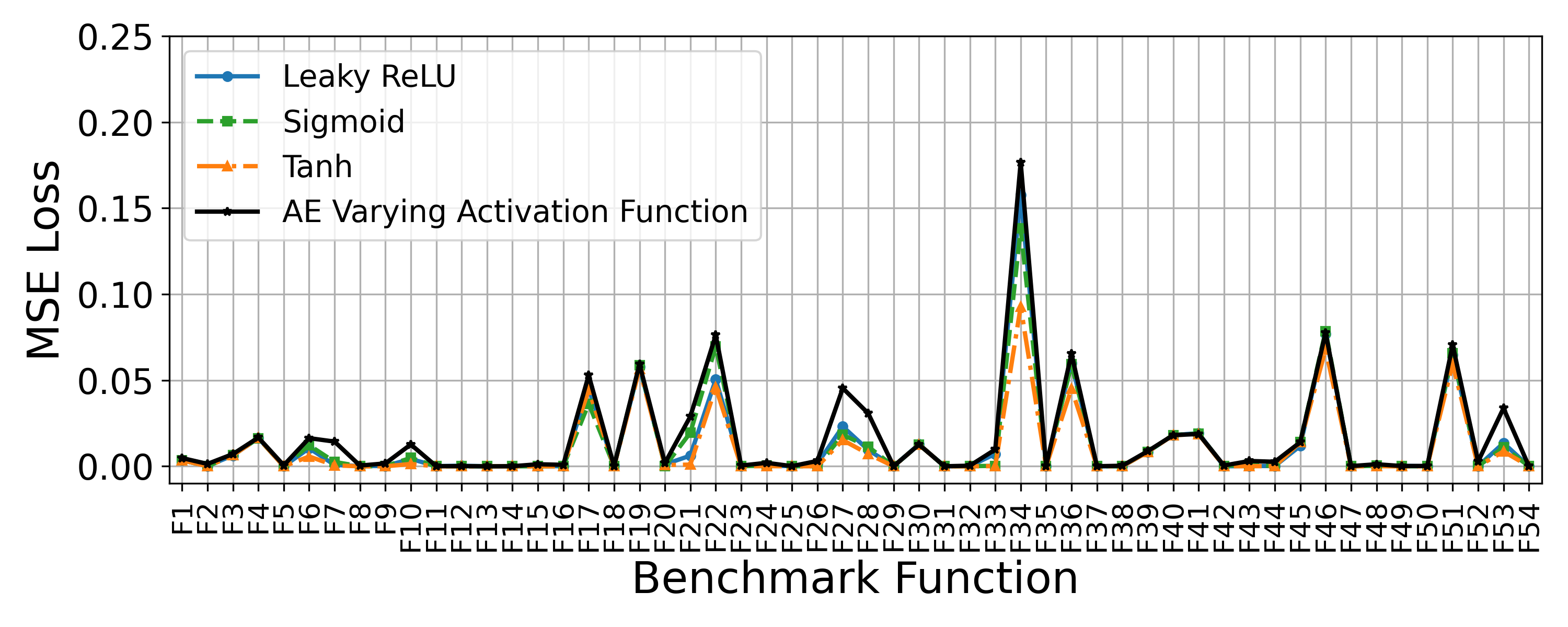}
    \caption{Comparison of the best MSEs obtained by traditional training baseline using Leaky ReLU, Sigmoid, and Tanh activations versus autoencoder-searched MLPs with neuron-level activation functions.}
    \label{fig:rare_act_compare}
\end{figure}

\autoref{fig:rare_act_compare} further compares our method to the traditional training baseline described in \autoref{Baselines}. Our method achieves performance comparable to MLPs trained using traditional training method with fixed activation functions in most cases. While the accuracy is similar or slightly lower, the ability to flexibly assign activation functions per neuron enables our method to produce significantly more compact and sparse networks, as demonstrated in \autoref{exp:all_components} and \autoref{Search for Sparse and Compact MLPs}.

\subsection{Search for Sparse MLPs}
\label{Search for Sparse and Compact MLPs}
Building upon the activation function search in \autoref{Search for Optimal MLPs with Activation Functions}, we compare our autoencoder-based search incorporated with sparsity penalties with the baseline of autoencoder-based search with ADMM afterwards described in \autoref{Baselines}. For our autoencoder-based method, we set $\lambda_s$ to $1 \times 10^{-3}$, and use $\mu_1 = 0.1$ for the $\ell_1$ penalty and $\mu_c = 0.01$ for the soft count term. Note that in this set of experiments, we do not apply neuron-level pruning yet.

\begin{figure}
    \centering
    \includegraphics[width=1\linewidth]{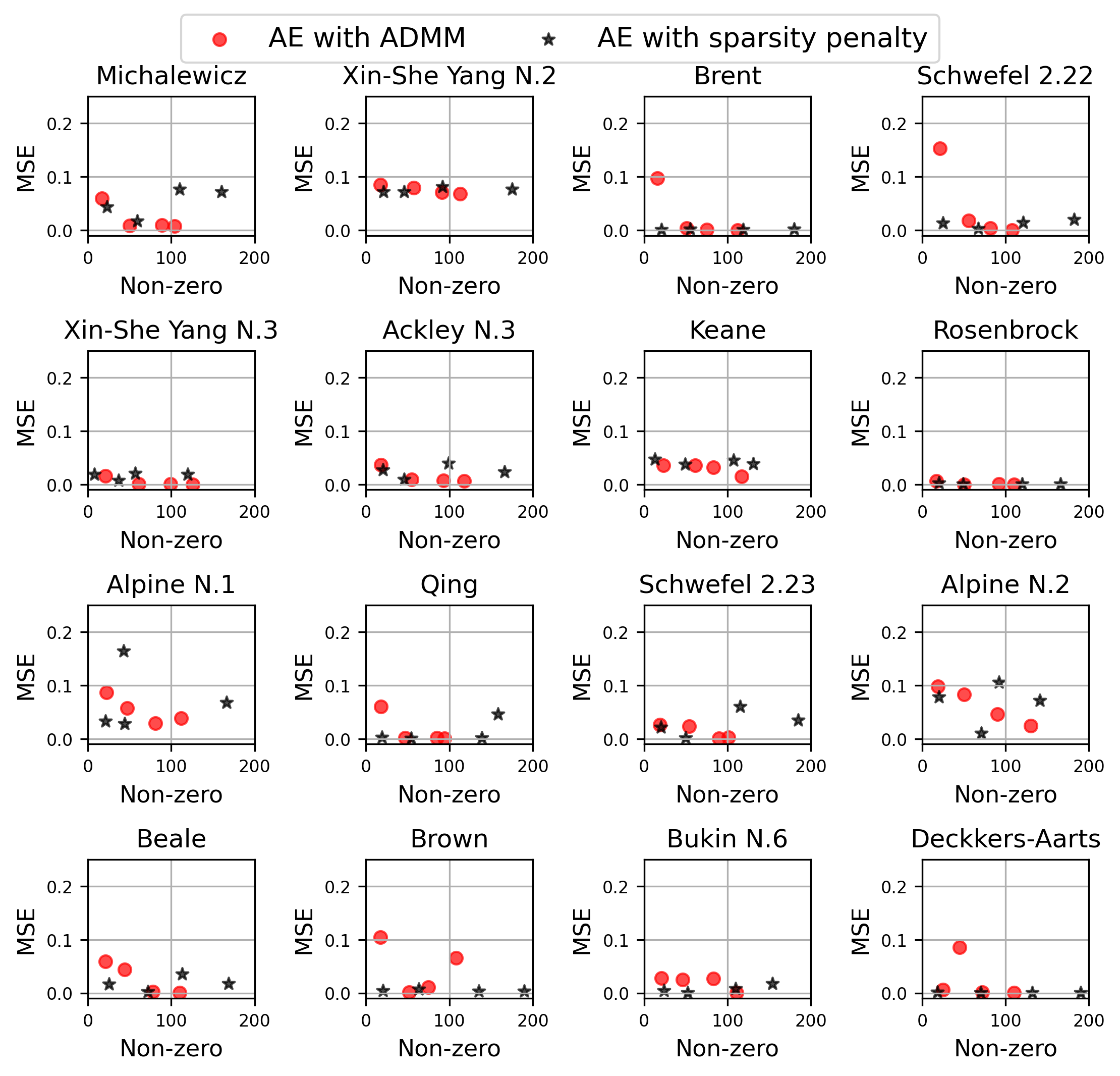}
    \caption{MSE vs. non-zero weights for autoencoder-based search with ADMM and with sparsity penalties. Each point represents a model configuration.}
    \label{fig:AE_ADMM_pareto}
\end{figure}

\begin{figure}
    \centering
    \includegraphics[width=1\linewidth]{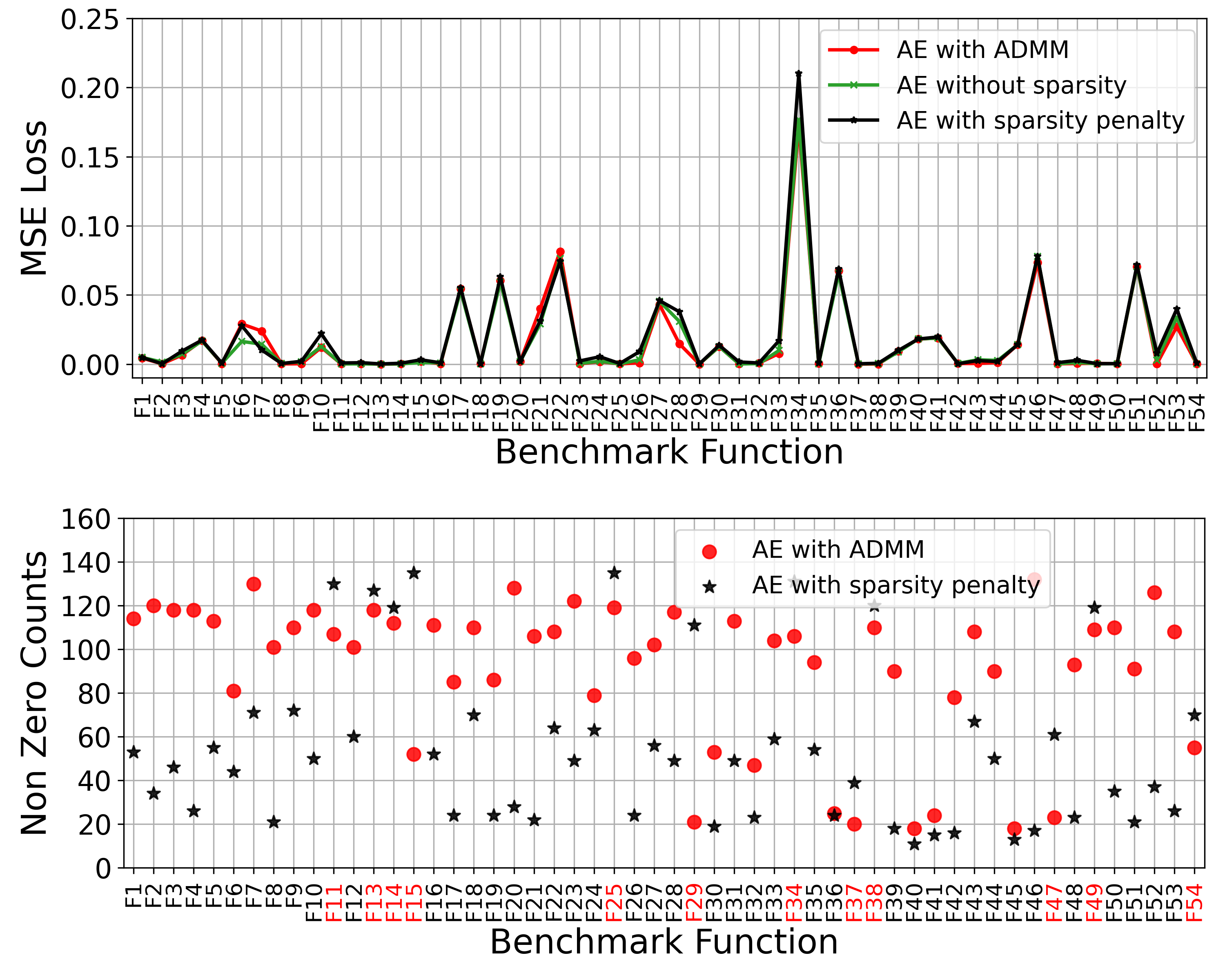}
    \caption{Top: MSE of the best-performing models, comparing AE followed by ADMM, AE followed by sparsity penalty, and AE without pruning. Bottom: Corresponding number of non-zero weights in the selected models. Function labels are color-coded to indicate which method yields a sparse model.}
    \label{fig:AE_ADMM_combine}
\end{figure}

The MSE vs non-zero weights scatter plots for 16 benchmark functions shown in \autoref{fig:AE_ADMM_pareto} indicates that our method consistently produces configurations that dominate or closely approach the left-down corner region. Meanwhile, \autoref{fig:AE_ADMM_combine} summarize the best selected model configurations in terms of MSE and number of non-zero weights. Compared to the models without any pruning, our autoencoder-based method with sparsity penalties achieves substantial reductions in non-zero weights while maintaining similar MSE. Furthermore, relative to the baseline where autoencoder-based search is followed by ADMM pruning, our simultaneous search consistently identifies more compact models. In most cases, it also achieves lower or comparable test MSE, demonstrating the effectiveness of integrating sparsity directly into the optimization process rather than decomposing architecture discovery and weight training or pruning into separate stages.

\section{Discussion}

\subsection{Smoothness of the Embedding Space} 
 
The embedding space learned by the multi-scale autoencoder is central to our framework, mapping functionally similar MLPs to nearby points in the embedding space. A smooth embedding space enables accurate gradient-based navigation toward optimal networks.
To assess smoothness of the embedding space for each decoder $D_k$, we sample a base embedding vector $z_{\text{init}} \sim \mathcal{N}(0, I)$ and generate 200 nearby embeddings by adding small Gaussian noise. We then apply principal component analysis (PCA) and extract the top two principal directions, denoted $v_1$ and $v_2$. These directions define a local 2D subspace around $z_{\text{init}}$, along which we generate perturbed embeddings $z{\prime} = z_{\text{init}} + \alpha v_1 + \beta v_2$, where $\alpha$, $\beta$ $\in [-3, 3]$ are sampled on a grid.

Each perturbed embedding $z{\prime}$, along with $z_{\text{init}}$, is decoded using $D_k$ to obtain the corresponding decoded MLPs. For a fixed set of input values, we compute the outputs of $D_k(z_{\text{init}})$ and $D_k(z{\prime})$, treating the former as ground truth. The MSE between the two outputs quantifies the functional deviation induced by perturbations in the embedding space.

\begin{figure}
    \centering
    \includegraphics[width=0.8\linewidth]{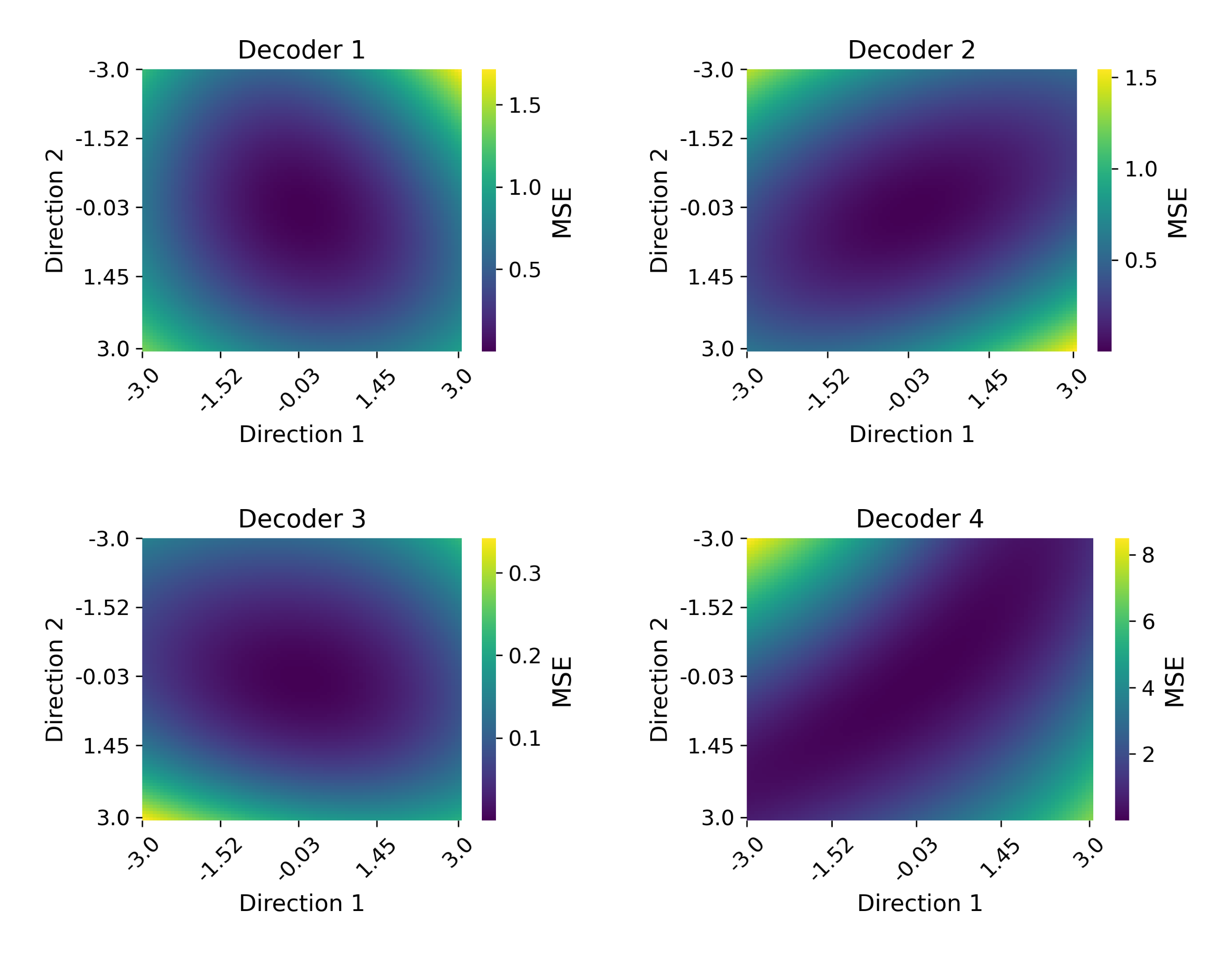}
    \caption{Visualization of latent space smoothness across 2D subspaces for the four decoders.}
    \label{fig:smoothness_2row}
\end{figure}

We visualize the resulting MSE values over the 2D grid of ($\alpha$, $\beta$) to assess the smoothness of the embedding space. \autoref{fig:smoothness_2row} shows that the MSE varies smoothly across directions $v_1$ and $v_2$, indicating that the latent space learned by the autoencoder preserves functional continuity.

\subsection{Comparison Between CNN-Based and GPT-Based Autoencoders}
\begin{figure}
    \centering
    \includegraphics[width=1\linewidth]{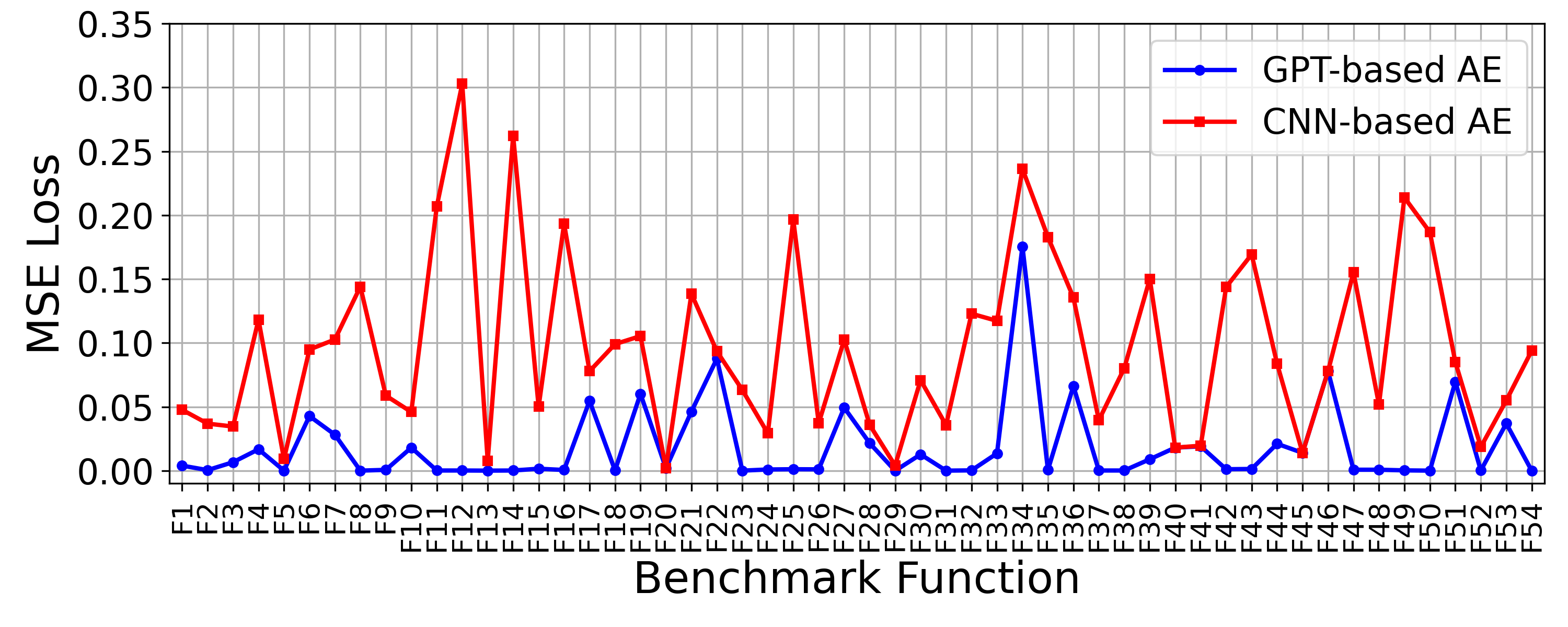}
    \caption{Comparison of minimum test MSE across benchmark function datasets using CNN-based and GPT-based autoencoders.}
    \label{fig:cnn_vs_gpt}
\end{figure}

As an ablation, we compare CNN-based and GPT-based autoencoders to examine which is more effective at capturing functional representations. Both autoencoders are trained under the same basic setting -- without bias terms, neuron masking, or activation function selection (all neurons use sigmoid). Using the same training configuration, we compare the best MSE achieved in downstream MLP search on CORNN benchmark suite. As shown in \autoref{fig:cnn_vs_gpt}, the GPT-based model consistently outperforms the CNN-based one.

\subsection{Functional Approximation by the Autoencoder}
\begin{figure}
    \centering
    \includegraphics[width=1\linewidth]{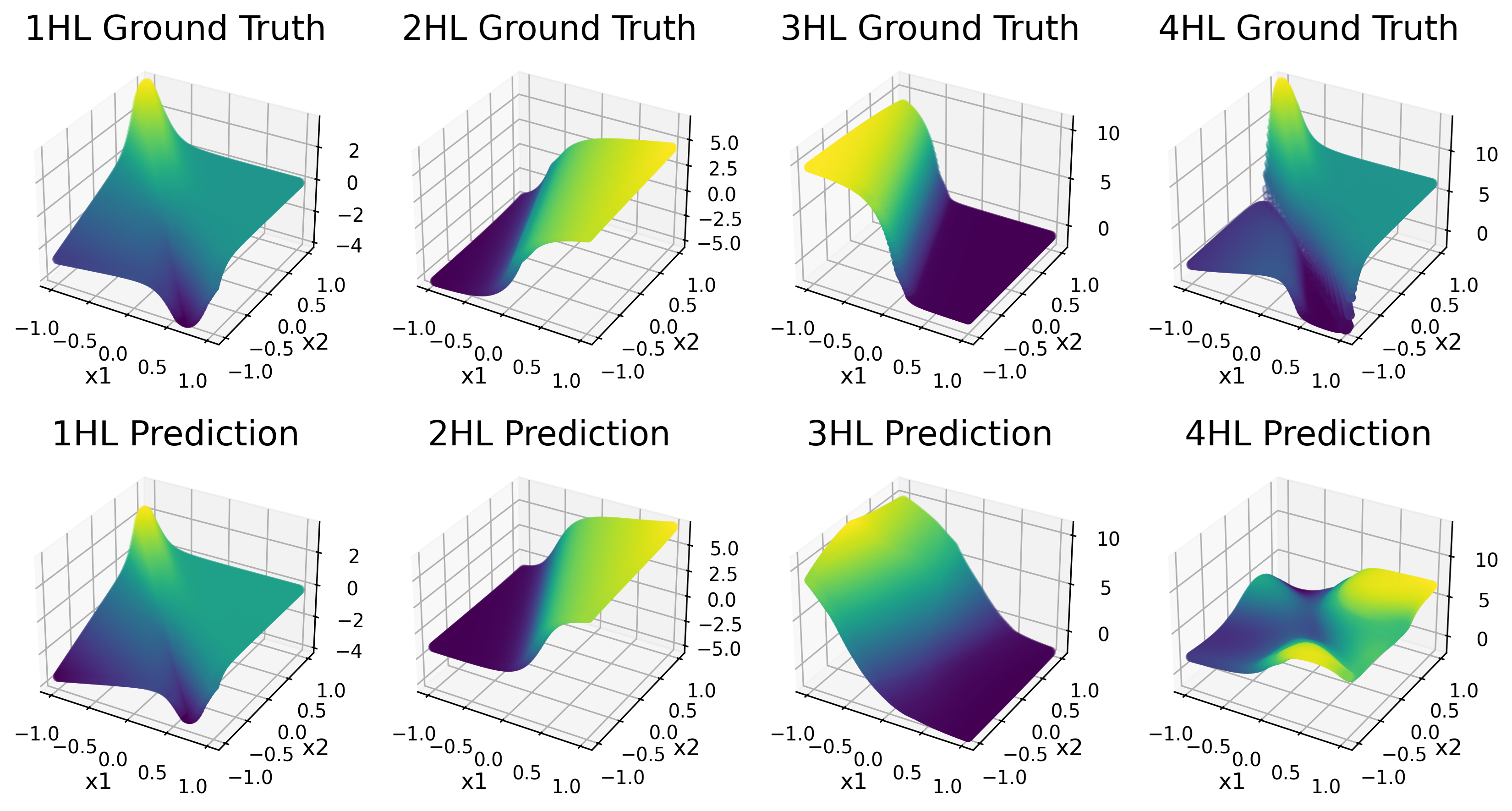}
    \caption{Comparison between the outputs of input MLPs and their best decoded counterparts  (minimum-MSE selection among all decoders). HL refers to hidden layers.}
    \label{fig:input_output}
\end{figure}

We visualize the decoded outputs corresponding to input MLPs with 1 to 4 hidden layers, as shown in \autoref{fig:input_output}. While the decoded MLPs generally approximate the functional behavior of the inputs, some discrepancies remain, particularly for deeper networks. However, as demonstrated in \autoref{all_exp}, our method is still able to discover high-performing MLPs through gradient-based optimization in the embedding space.

We hypothesize that the embedding space only needs to provide a sufficiently smooth and semantically organized landscape, such that gradient descent can navigate toward better-performing regions. It does not need to achieve perfect reconstruction, but rather serve as a functional embedding mechanism. Improving the fidelity of the autoencoder would enhance the smoothness and precision of the search space, leading to better performance in the downstream optimization.

\subsection{Different hyperparameter choices}
In \autoref{eq:overall_sp_loss}, \autoref{eq:sp_loss}, and \autoref{eq:compact_loss}, the sparsity and compactness behavior of the searched neural networks is primarily governed by three key hyperparameters: $\lambda_s$ for controlling the overall strength of sparsity, and the internal weighting coefficients $\alpha$ and $\beta$ used in the neuron compactness term. We evaluate three settings with increasing regularization strength, as summarized in \autoref{tab:penalty_settings}.

\begin{table}[h]
\centering
\caption{Hyperparameter settings for different levels of sparsity and compactness regularization.}
\label{tab:penalty_settings}
\begin{tabular}{lccc}
\toprule
\textbf{Penalty Level} & $\lambda_s$ & $\alpha$ & $\beta$ \\
\midrule
Small Penalty  & $1 \times 10^{-5}$ & $1 \times 10^{-1}$ & $1 \times 10^{-4}$ \\
Medium Penalty & $1 \times 10^{-4}$ & $4 \times 10^{-1}$ & $1 \times 10^{-3}$ \\
Large Penalty  & $1 \times 10^{-3}$ & $4 \times 10^{-1}$ & $1 \times 10^{-1}$ \\
\bottomrule
\end{tabular}
\end{table}

\begin{figure}
    \centering
    \includegraphics[width=1\linewidth]{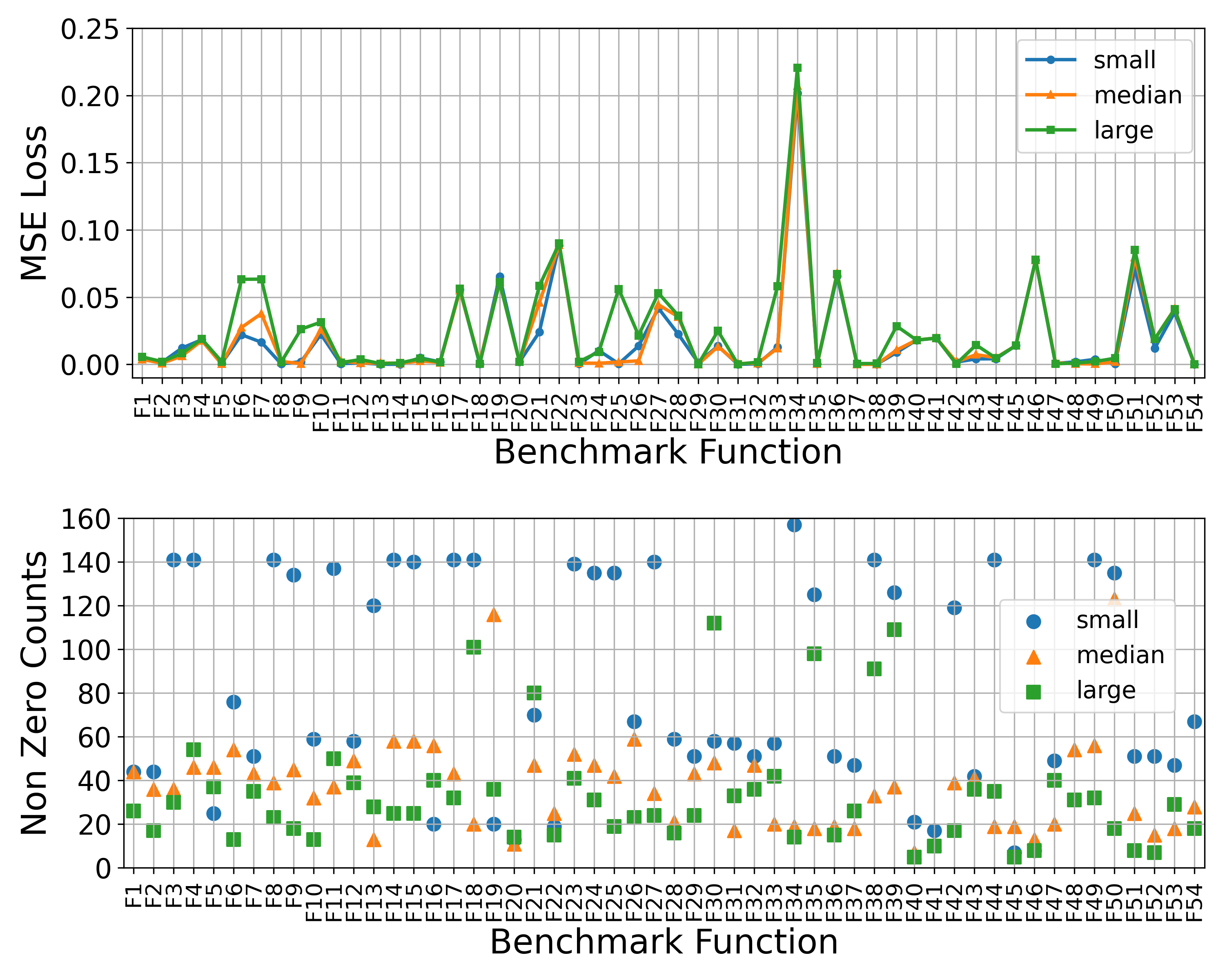}
    \caption{Test MSE (top) and non-zero weights (bottom) for the best networks under three sparsity and compactness regularization levels.}
    \label{fig:Hyper}
\end{figure}

\autoref{fig:Hyper} illustrates the effect of different regularization settings on the performance and sparsity of the resulting models. As expected, the curves align well with the penalty levels defined in \autoref{tab:penalty_settings}: stronger regularization (i.e., larger values of $\lambda_s$, $\alpha$, and $\beta$) generally leads to lower non-zero weight counts but higher MSE. This confirms the trade-off between model compactness and predictive accuracy. In practice, medium penalty settings tend to offer a favorable balance, yielding compact architectures with minimal loss in performance.

\subsection{Change in size of the embedding space}
\begin{figure}
    \centering
    \includegraphics[width=1\linewidth]{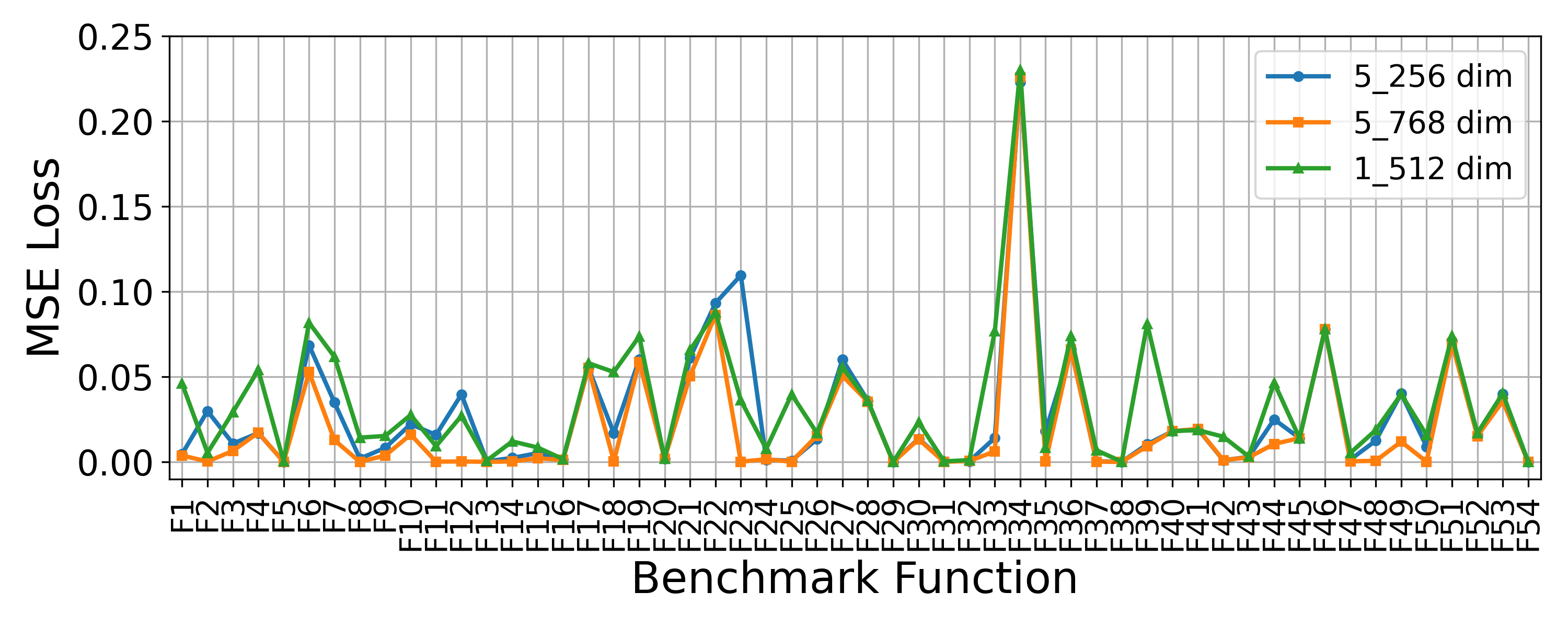}
    \caption{Test MSE for the best MLPs found using embedding vectors of different dimensions on the CORNN benchmark suite. Each curve corresponds to a specific embedding size: $5 \times 768$, $5 \times 256$, and $1 \times 512$.}
     \label{fig:latent_dim}
\end{figure}

To assess whether a large embedding dimensionality is necessary, we train autoencoders to embed 1–4 hidden layer MLPs with fixed structure: no bias terms, no activation functions, and exactly 5 neurons per layer. Each autoencoder is trained for 50 epochs, with 640{,}000 sampled MLPs per epoch. \autoref{fig:latent_dim} compares performance on all regression datasets across different embedding sizes. In nearly all datasets, the $5 \times 768$ embedding yields lower or equal MSE loss compared to smaller alternatives ($5 \times 256$ and $1 \times 512$). This suggests that a sufficiently large embedding space is beneficial for capturing functional variations in MLPs. Accordingly, we use a $7 \times 768$ embedding in our main experiments.

\subsection{Compression of networks and variable input/output size}
\label{disc:compression}

\begin{figure*}
    \centering
    \includegraphics[width=0.7\linewidth]{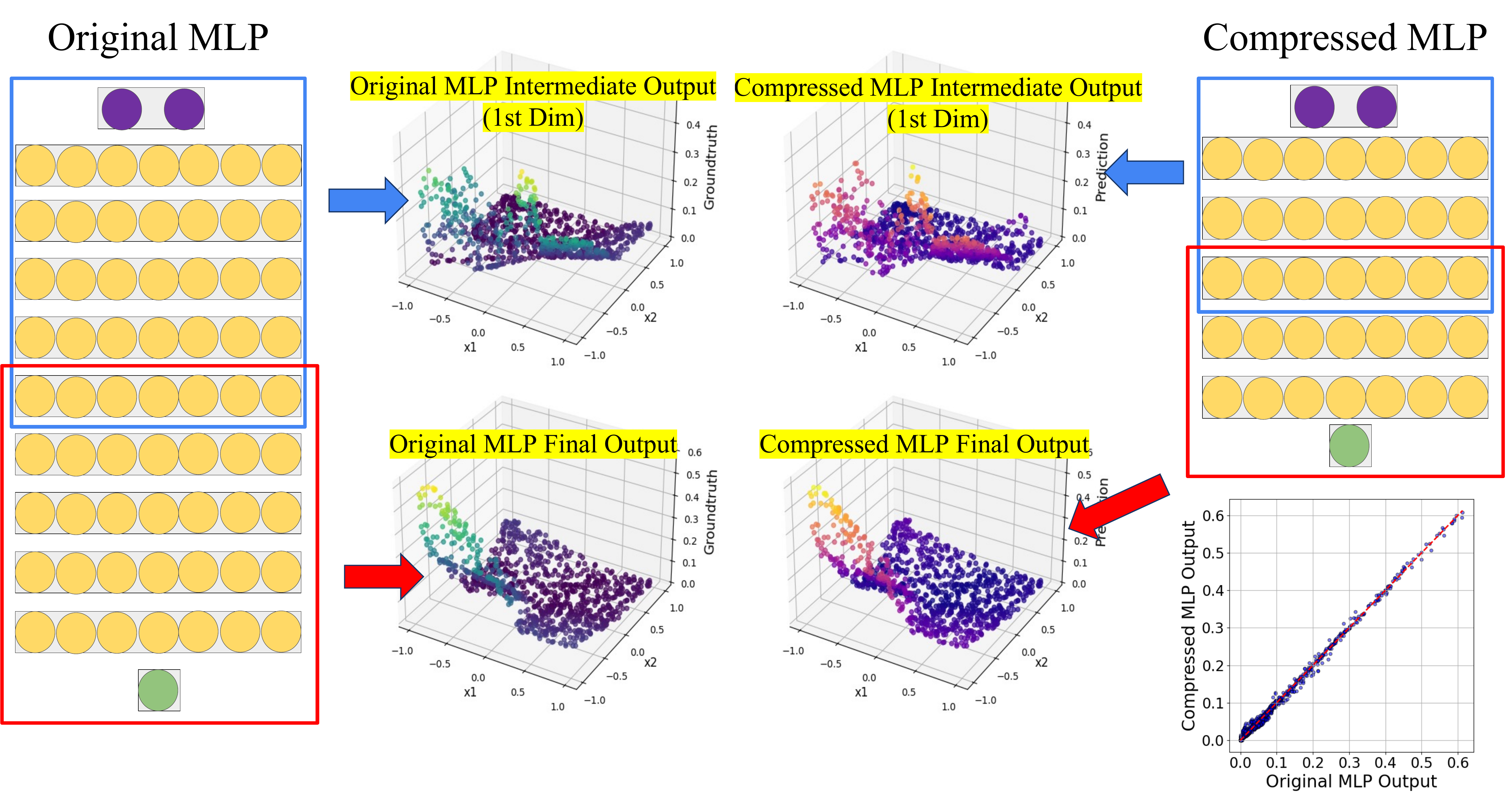}
    \caption{Intermediate and final outputs of the original vs. compressed MLP. The compressed 9 hidden layer MLP closely replicates both internal representations and final predictions of the original 9 hidden layer MLP. This approach also demonstrates SWAT-NN's ability to train variable input/output MLPs.}
    \label{fig:compression_plot}
\end{figure*}

It is possible to compress large MLPs into smaller ones using the existing embedding space, without requiring deeper autoencoders. The key idea is that a deep network can be decomposed into smaller sub-networks connected in sequence, each of which can be individually compressed using the autoencoder. As a proof of concept, we compress a 9-hidden-layer MLP into a 5-hidden-layer MLP by splitting it into two sub-networks and compressing each with a lightweight 4-to-2 autoencoder. This autoencoder is trained on MLPs with leaky ReLU activations, 7 neurons per layer, and no biases.

Importantly, our matrix-based representation supports varying input and output sizes by updating the second-channel mask (see \autoref{fig:matrix_repr}). This flexibility allows us to compress the first 4-layer MLP with input size 2 and output size 7, and the second with input size 7 and output size 1.

The compression is achieved by optimizing in the latent space of the 4-to-2 autoencoder for each sub-network. As shown in \autoref{fig:compression_plot}, the output of the compressed 5-layer MLP closely matches that of the original 9-layer MLP, demonstrating that large networks can be effectively approximated using SWAT-NN without needing to retrain a deeper autoencoder that matches the size of the original large network.

\section{Conclusion}
We presented SWAT-NN, a novel framework that simultaneously optimizes both neural architectures and their weights within a universal and continuous embedding space. Through extensive experiments on all 54 regression datasets in the CORNN benchmark suite, we demonstrated that SWAT-NN consistently yields performant networks with desirable sparsity and compactness, achieving comparable or better accuracy while producing significantly more compact and sparse neural networks than state-of-the-art methods.

While this work focuses on MLPs, the SWAT-NN framework opens up several future research directions. In particular, extending the approach to more complex neural network families such as Temporal Convolutional Networks (TCNs), Convolutional Neural Networks (CNNs), and Recurrent Neural Networks (RNNs) could enable broader applicability.

\bibliographystyle{IEEEtran}
\bibliography{reference}

\end{document}